\documentclass[journal]{IEEEtran}

\usepackage[utf8]{inputenc}
\usepackage[T1]{fontenc}
\usepackage{times}
\usepackage{amsmath,amssymb,amsthm}
\usepackage{graphicx}
\usepackage{booktabs}
\usepackage{multirow}
\usepackage{algorithm}
\usepackage{algorithmic}
\usepackage{caption}
\usepackage{subcaption}
\usepackage{hyperref}
\usepackage{cite}
\usepackage{xcolor}
\usepackage[table]{xcolor}
\usepackage{tikz}
\usepackage{pgfplots}
\pgfplotsset{compat=1.18}
\usepackage{stfloats}

\hypersetup{
  colorlinks=true,
  linkcolor=black,
  citecolor=blue,
  urlcolor=blue
}

\title{AquaFusionNet: Lightweight Vision--Sensor Fusion Framework for Real-Time Pathogen Detection and Water Quality Anomaly 
Prediction on Edge Devices}

\author{
  \IEEEauthorblockN{
  Sepyan~Purnama~Kristanto,
  Lutfi~Hakim, and
  Hermansyah%
  \thanks{S.~P.~Kristanto and L.~Hakim are with the Department of Informatics Engineering, Politeknik Negeri Banyuwangi, Banyuwangi, Indonesia (e-mail: sepyan@poliwangi.ac.id).}%
  \thanks{Hermansyah is with Balai Besar Teknik Kesehatan Lingkungan dan P2B Surabaya, Indonesia.}%
  }
}

\begin{document}

\maketitle

\begin{abstract}
Evidence from many low- and middle-income regions shows that microbial contamination in small-scale drinking-water systems often fluctuates rapidly, yet existing monitoring tools capture only fragments of this behaviour. Microscopic imaging provides organism-level visibility, whereas physicochemical sensors reveal short-term changes in water chemistry; in practice, operators must interpret these streams separately, making real-time decision-making unreliable.

This study introduces \textbf{AquaFusionNet}, a lightweight cross-modal framework that unifies both information sources inside a single edge-deployable model. Unlike prior work that treats microscopic detection and water-quality prediction as independent tasks, AquaFusionNet learns the statistical dependencies between microbial appearance and concurrent sensor dynamics through a gated cross-attention mechanism designed specifically for low-power hardware. The framework is trained on \textbf{AquaMicro12K}, a new dataset comprising 12{,}846 annotated 1000$\times$ micrographs curated for drinking-water contexts, an area where publicly accessible microscopic datasets are scarce.

Deployed for six months across seven facilities in East Java, Indonesia, the system processed 1.84~million frames and consistently detected contamination events with 94.8\% mAP@0.5 and 96.3\% anomaly-prediction accuracy, while operating at 4.8~W on a Jetson Nano. Comparative experiments against representative lightweight detectors show that AquaFusionNet provides higher accuracy at comparable or lower power, and field results indicate that cross-modal coupling reduces common failure modes of unimodal detectors, particularly under fouling, turbidity spikes, and inconsistent illumination. All models, data, and hardware designs are released openly to facilitate replication and adaptation in decentralized water-safety infrastructures.
\end{abstract}

\begin{IEEEkeywords}
Edge AI, multimodal fusion, water quality, pathogen detection, lightweight CNN, Internet of Things.
\end{IEEEkeywords}

\IEEEpeerreviewmaketitle

\section{Introduction}
\label{sec:intro}

Safe drinking water is a prerequisite for public health, yet it remains out of reach for a substantial fraction of the global population. Recent estimates from the WHO/UNICEF Joint Monitoring Programme indicate that 2.2~billion people still lack safely managed drinking-water services and that unsafe water, sanitation, and hygiene (WASH) contribute to approximately 1.4~million deaths per year~\cite{who_unicef_2021,wolf_2023_burden}. The burden is concentrated in low- and middle-income countries, where waterborne diarrhoeal disease continues to be a leading cause of morbidity and mortality among children under five.

In many such settings, households obtain drinking water not from centralized utilities but from small-scale depots, community-managed treatment units, and informal vendors. These systems are attractive because they are flexible and locally managed, but they also introduce new risks. Source quality can vary substantially over short time scales, equipment is often operated with minimal training, and maintenance is reactive rather than preventive. Field investigations regularly report exceedances of microbiological standards in these systems, despite nominal compliance with basic treatment requirements.

Routine monitoring remains dominated by culture-based microbiological assays performed in centralized laboratories. While such tests are sensitive and well understood, they are slow: typical turn-around times range from 24 to 48~hours or longer. During this delay, water that is later found to be contaminated may have already been distributed and consumed. Operators thus require additional, faster feedback channels if they are to take meaningful action during contamination events.

Two streams of technology have emerged to fill this gap. First, low-cost physicochemical sensors---for turbidity, pH, conductivity, dissolved oxygen, and related parameters---are now widely used in IoT-based monitoring systems~\cite{ren_2022_edgewater,ajakwe_2023_ciswqms,ngwenya_2025_ml_iot}. These sensors can provide continuous measurements and threshold-based alarms with modest power requirements. Second, compact digital microscopy, coupled with machine learning, has been explored as a means to automatically detect bacteria, algae, and other contaminants in micrographs or agar-plate images~\cite{yanik_2020_ecoli,sonmez_2023_microalgae,ramadhani_2024_yolov8microalgae}. Each of these modalities captures a different facet of water quality: sensors describe the chemical and physical environment, while microscopy reveals the organisms and particles themselves.

In practice, however, these streams are almost always interpreted separately. Operators may observe a turbidity spike and suspect contamination, but without direct visualization they lack confirmation. Conversely, a microscope image may show bacterial colonies, yet the broader chemical context is missing. This separation is not merely operational; it is reflected in the research landscape. Most deep-learning models developed to date are either vision-based detectors trained on laboratory imagery, or time-series models that infer aggregate indices from sensor data~\cite{im_2022_tapwater,ehteram_2024_hybridwqi,helaly_2025_mldlreview}. To the best of our knowledge, there is no existing model that learns how microbial appearance and short-term sensor dynamics co-evolve in real small-scale systems.

\subsection{Problem Statement and Motivation}
\label{subsec:problem}

The motivating observation for this work is that, in field practice, operators implicitly think in cross-modal terms. Interviews with depot owners and technicians reveal recurring narratives: sudden spikes in turbidity, changes in oxidation--reduction potential (ORP), or temperature shifts that coincide with visible flocs, microalgae blooms, or microplastics under the microscope. These relationships are qualitatively recognized but not formalized in existing monitoring tools.

We therefore ask: \emph{Can we design a lightweight, edge-deployable neural model that (i) processes microscopic images and multivariate sensor streams jointly, and (ii) improves the reliability of contamination alerts under realistic noise and resource constraints?} This question is not purely algorithmic; it is also constrained by hardware, cost, and operational realities. Edge devices deployed in rural depots must be robust, low-power, and inexpensive, with limited connectivity.

\subsection{Limitations of Existing Approaches}
\label{subsec:limitations}

From a modelling perspective, three limitations stand out:

\begin{itemize}
  \item \textbf{Unimodal reasoning.} Vision-only models cannot anticipate contamination that is chemically detectable before it is visually apparent, while sensor-only models cannot exploit the detailed spatial structure present in micrographs. Either modality alone is therefore vulnerable to specific failure modes.
  \item \textbf{Resource mismatch.} Multimodal fusion models in the broader environmental AI literature frequently assume datacenter or high-end GPU resources~\cite{gao_2020_multimodal,jiao_2024_multimodal}. Their computational footprint, memory usage, and latency are incompatible with low-cost edge devices that must run continuously on solar power.
  \item \textbf{Lack of field validation.} Many proposed architectures are evaluated exclusively on static benchmark datasets. It remains unclear how such models behave under long-term deployment with sensor fouling, lighting drift, and shifting source-water regimes.
\end{itemize}

These limitations suggest that simply porting existing multimodal architectures to an edge platform is unlikely to be sufficient. Instead, we require a design that is explicitly co-optimized for both algorithmic performance and hardware constraints.

\subsection{Contributions}
\label{subsec:contributions}

This paper makes the following contributions:

\begin{enumerate}
  \item \textbf{Problem formulation.} We formulate drinking-water safety monitoring as a cross-modal prediction task that couples microscopic imagery with short-horizon physicochemical sensor data. To our knowledge, this formulation has not previously been explored in the water-quality literature.
  \item \textbf{AquaMicro12K dataset.} We introduce \emph{AquaMicro12K}, a microscopic contaminant dataset comprising 12{,}846 annotated 1000$\times$ micrographs from drinking-water contexts, covering eight classes including bacteria, protozoa, algae, microplastics, and inorganic particles. The dataset is curated to reflect conditions in small-scale depots rather than laboratory plates.
  \item \textbf{AquaFusionNet architecture.} We propose \emph{AquaFusionNet}, a hardware-aware fusion architecture that combines a pruned SSD--MobileNetV3-Small vision branch with a compact temporal CNN (AquaTemp-Net) and a gated cross-attention module. The fusion block is designed to add only 0.4\,M parameters while providing robustness to noisy sensor readings.
  \item \textbf{Edge-optimized implementation.} We jointly optimize the model and deployment stack for an NVIDIA Jetson Nano--based platform augmented by an ESP32-S3 microcontroller and Raspberry Pi HQ Camera. Structured pruning and INT8 quantization are applied under a performance budget of 8.7\,MB, 4.8\,W, and 41\,FPS.
  \item \textbf{Field validation.} We present results from a six-month deployment of seven AquaFusionNet nodes in East Java, Indonesia, analysing 1.84\,million processed frames. We report false positive and false negative rates against laboratory confirmation and characterise how cross-modal fusion alters the error profile relative to unimodal baselines.
\end{enumerate}

The rest of the paper is structured as follows. Section~\ref{sec:related} reviews related work in IoT-based water quality monitoring, deep learning for water quality, microscopic detection, and multimodal fusion. Section~\ref{sec:methods} describes the deployment setup, dataset, architecture, and training procedure. Section~\ref{sec:results} presents quantitative and qualitative results, including ablations and field analysis. Section~\ref{sec:discussion} discusses implications, limitations, and ethical considerations, and Section~\ref{sec:conclusion} concludes.

\section{Related Work}
\label{sec:related}

We situate AquaFusionNet at the intersection of four research areas: IoT and edge-based water quality monitoring, deep learning for water quality prediction, microscopic detection of pathogens and algae, and multimodal fusion on resource-constrained devices.

\subsection{IoT and Edge-Based Water Quality Monitoring}

A substantial body of work has explored architectures that combine low-cost sensors, communication modules, and cloud platforms for real-time water quality monitoring. Early systems typically employed microcontrollers with Wi-Fi or cellular links to stream data such as pH, turbidity, total dissolved solids (TDS), and temperature to remote dashboards~\cite{ren_2022_edgewater,ajakwe_2023_ciswqms}. More recent designs incorporate local processing to reduce bandwidth and latency, as well as to support simple rule-based control.

Ren \emph{et al.}~\cite{ren_2022_edgewater} propose an edge-computing-based online monitoring method that partitions computation between local nodes and the cloud to improve responsiveness and scalability. The CIS-WQMS architecture~\cite{ajakwe_2023_ciswqms} integrates connected intelligence, IoT sensors, and machine learning to track water distribution networks. Ngwenya \emph{et al.}~\cite{ngwenya_2025_ml_iot} provide a recent review of machine-learning-based approaches for IoT-enabled water quality monitoring, emphasising the importance of resource-aware models for deployment in constrained environments. Other works target specific domains such as wastewater treatment, aquaculture, and groundwater management~\cite{forhad_2024_iotwtp,nguyen_2024_aquaculture,maher_2023_wqa}.

These studies demonstrate that IoT-based monitoring can significantly improve spatial and temporal coverage, reduce manual sampling burdens, and enable basic early-warning functionality. However, most systems are sensor-only: they typically do not incorporate microscopic imagery, and machine learning components are often limited to regression or classification over physicochemical time series. Microscopy, when performed, tends to remain in the laboratory.

\subsection{Deep Learning for Water Quality Prediction}

Deep learning methods have been applied to a range of water-quality prediction problems, including forecasting indices, estimating parameters from remote sensing, and assisting laboratory workflows. Im \emph{et al.}~\cite{im_2022_tapwater} combine convolutional and recurrent networks to predict tap water quality from turbidity, chlorine residual, and pH measurements, reporting better temporal modelling than traditional baselines. Ehteram \emph{et al.}~\cite{ehteram_2024_hybridwqi} develop a hybrid CNN--clockwork RNN architecture to estimate a water quality index, illustrating the value of multi-scale temporal features.

Peterson \emph{et al.}~\cite{peterson_2020_algae} use deep networks on satellite-derived water colour to estimate blue-green algae and chlorophyll-$\alpha$ concentrations, bridging remote sensing and in-situ monitoring. Zhi \emph{et al.}~\cite{zhi_2024_dlwater} discuss how deep learning can accelerate laboratory-based assessments, particularly in microbiology. Review articles~\cite{helaly_2025_mldlreview,yan_2024_mlwqreview} synthesise these developments, pointing out common challenges such as data scarcity, non-stationarity, and interpretability.

Most of these models, however, remain unimodal. They work either on sensor time series, remotely sensed imagery, or laboratory datasets; they do not attempt to link pixel-level microscopic appearance with the concurrent physicochemical environment of the water sample. The multimodal formulation we adopt here is complementary to this body of work.

\subsection{Microscopic Pathogen and Algae Detection}

Digital microscopy coupled with convolutional neural networks has shown promise in automating pathogen detection in water-related samples. Yanik \emph{et al.}~\cite{yanik_2020_ecoli} employ Faster R-CNN to detect \emph{Escherichia coli} colonies on agar plates for rapid assessment. Other studies apply CNN-based pipelines to bacterial detection in drinking water and wastewater~\cite{khan_2021_ecoli,irani_2021_ecoli}.

For microalgae, S\"{o}nmez \emph{et al.}~\cite{sonmez_2023_microalgae} and others report high-accuracy classification using CNNs on light and electron microscopy images, while Ramadhani \emph{et al.}~\cite{ramadhani_2024_yolov8microalgae} employ YOLOv8-based models to detect \emph{Spirulina platensis} and \emph{Chlorella vulgaris}. Hussin \emph{et al.}~\cite{hussin_2025_microalgae} present a freshwater microalgae dataset with extensive labels to support further work in this area.

Despite their strengths, these systems are usually evaluated on curated laboratory datasets and assume either stationary imaging conditions or lab-prepared samples. Few studies consider deployment on low-power edge hardware, and none, to our knowledge, combine microscopic imagery with real-time physicochemical measurements in a single predictive model.

\subsection{Multimodal Fusion and Lightweight Edge Models}

Multimodal fusion techniques aim to integrate heterogeneous data sources such as images, text, audio, and time series. Surveys by Gao \emph{et al.}~\cite{gao_2020_multimodal} and Jiao \emph{et al.}~\cite{jiao_2024_multimodal} categorise fusion approaches into early, late, and hybrid strategies, and highlight the increasingly important role of attention mechanisms in learning cross-modal correspondences.

Within environmental monitoring and industrial anomaly detection, attention-based fusion has been applied to hyperspectral and LiDAR data~\cite{mohla_2020_fusatnet}, multimodal temporal signals~\cite{qu_2024_mfgan,zhou_2024_mmanomaly}, and sensor networks~\cite{wang_2023_mefn}. On the architectural side, lightweight backbones such as MobileNetV3~\cite{howard_2019_mobilenetv3} and one-stage detectors such as SSD, YOLOv5, and YOLOv8~\cite{liu_2016_ssd,fan_2024_yolov8_underwater} enable real-time vision on embedded platforms, including NVIDIA Jetson Nano and ARM SoCs. Detection transformers like RT-DETR provide improved accuracy at the expense of additional complexity~\cite{rtdetr_2024}.

Edge computing paradigms emphasise pushing inference closer to the data source to reduce latency and bandwidth usage~\cite{shi_2016_edge}. However, most multimodal architectures have been designed with server-class hardware in mind. There is limited guidance on how to design fusion mechanisms that explicitly respect the constraints of low-cost edge devices. AquaFusionNet can be viewed as an attempt to bring multimodal attention into the edge-computing regime for a concrete, socially relevant application.

\section{Materials and Methods}
\label{sec:methods}

In this section we describe the field deployment, the AquaMicro12K dataset, the AquaFusionNet architecture, the training procedure, and the statistical analysis used to compare models.

\subsection{Field Deployment}
\label{subsec:deployment}

From January to June 2025, we deployed seven AquaFusionNet edge nodes in Banyuwangi Regency, East Java, Indonesia: five at refill-style drinking-water depots and two at river intake points supplying small-scale treatment systems. Sites were selected in collaboration with local health authorities to represent a mix of urban and peri-urban contexts, and to capture different source-water regimes (groundwater, surface water, and mixed).

Each node consists of:

\begin{itemize}
  \item A custom 1000$\times$ digital microscope based on a Raspberry Pi HQ Camera with a 1000$\times$ achromatic objective and LED illumination, enclosed in a dust- and splash-resistant housing.
  \item Commercial sensors for pH, turbidity, TDS, temperature, dissolved oxygen (DO), and oxidation--reduction potential (ORP), connected via I\(^2\)C and UART to an ESP32-S3 microcontroller.
  \item An NVIDIA Jetson Nano 4\,GB module handling image capture, neural inference, logging, and communication with a remote server over 4G.
  \item A 20\,W solar panel and a 12\,Ah LiFePO\(_4\) battery providing autonomous power, with automatic switchover to grid power where available.
\end{itemize}

Water is drawn from the source through a pre-filter and pumped through a sensor manifold and a microscopic imaging chamber before discharge. The manifold ensures that all sensors and the microscope view essentially the same water volume, albeit with small temporal offsets. The ESP32-S3 samples sensors at 1\,Hz, aggregates a 60-second window, and transmits it to the Jetson via UART. The Jetson triggers the camera at 1--2\,FPS, performs inference with AquaFusionNet, logs detections and risk scores, and sends periodic summaries to a remote server.

\begin{figure}[t]
  \centering
  \includegraphics[width=0.95\columnwidth]{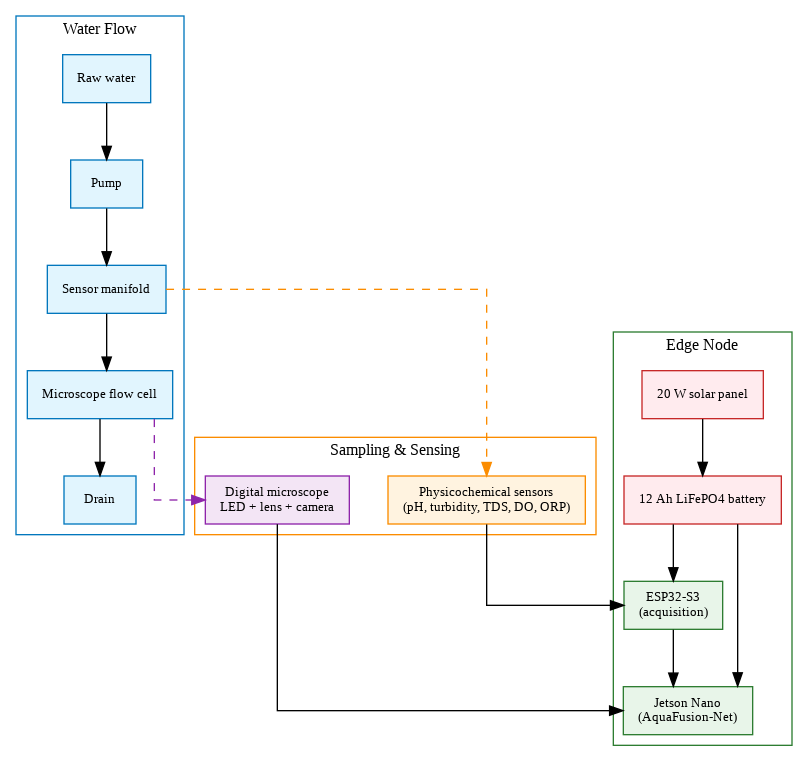}
  \caption{Schematic of an AquaFusionNet deployment at a drinking-water depot, showing the water source, pump, sensor manifold, microscopic imaging chamber, edge compute node (ESP32-S3 + Jetson Nano), and solar-powered supply. Water flows through the manifold and microscope chamber before discharge, while sensor and image streams are processed locally on the edge node.}
  \label{fig:deployment}
\end{figure}

Figure~\ref{fig:deployment} sketches the hydraulic, sensing, imaging, compute, and power subsystems at a typical depot installation.

\subsection{AquaMicro12K Dataset}
\label{subsec:dataset}

AquaMicro12K comprises 12{,}846 RGB microscopic images captured at 1000$\times$ magnification from raw and treated water samples collected during deployment and supplementary sampling campaigns. Samples were taken across dry and rainy seasons to capture a range of conditions, including high-turbidity events and periods of relative stability.

Images were annotated independently by three certified microbiologists using a custom web-based tool. Annotators drew bounding boxes around visible targets and assigned class labels from a predefined taxonomy. Consensus labels were obtained via majority voting, with disagreements resolved through a short adjudication session. Inter-annotator agreement measured by Fleiss' $\kappa$ was 0.89, indicating near-perfect agreement.

The eight object classes are:
\begin{itemize}
  \item \emph{E.~coli} (rod-shaped bacteria),
  \item Total coliform,
  \item \emph{Pseudomonas aeruginosa},
  \item \emph{Enterococcus},
  \item \emph{Giardia lamblia} (cysts and trophozoites),
  \item Microplastics (fibres and fragments),
  \item Algae (including filamentous and unicellular forms),
  \item Inorganic particles (e.g., silt, rust).
\end{itemize}

Table~\ref{tab:dataset} summarises the per-class image counts and average object density. Some classes, such as \emph{Giardia}, are rare, while particulate classes are more common and dense. This class imbalance reflects real-world conditions and poses a challenge for detection models.

\begin{table}[h]
  \centering
  \caption{AquaMicro12K dataset statistics.}
  \begin{tabular}{lrr}
  \toprule
  Class                  & Images & Avg. objects/image \\
  \midrule
  E. coli                & 2,843 & 6.4 \\
  Total coliform         & 2,156 & 5.1 \\
  Pseudomonas aeruginosa & 1,789 & 4.8 \\
  Enterococcus           & 1,412 & 3.9 \\
  Giardia lamblia        &   987 & 1.2 \\
  Microplastics          & 1,234 & 8.2 \\
  Algae                  & 1,105 & 12.1 \\
  Inorganic particles    & 1,320 & 15.3 \\
  \midrule
  Total                  & 12,846 & 7.1 \\
  \bottomrule
  \end{tabular}
  \label{tab:dataset}
\end{table}

To avoid leakage, we split images into training (70\%), validation (15\%), and test (15\%) sets at the sample level; all images derived from a given physical sample are assigned to the same split. Data augmentation includes random horizontal and vertical flips, random crops, mild Gaussian blur, and colour jitter in HSV space. Images are resized to 416$\times$416 pixels for training and inference, which we found to offer a good trade-off between detail and throughput on Jetson Nano.

\subsection{AquaFusionNet Architecture}
\label{subsec:architecture}

AquaFusionNet comprises three main components: (i) a vision branch for microscopic detection, (ii) a temporal branch (AquaTemp-Net) for sensor time series, and (iii) a gated cross-attention module that fuses features from both branches.

\begin{figure*}[!t]
  \centering
  \includegraphics[width=0.92\textwidth]{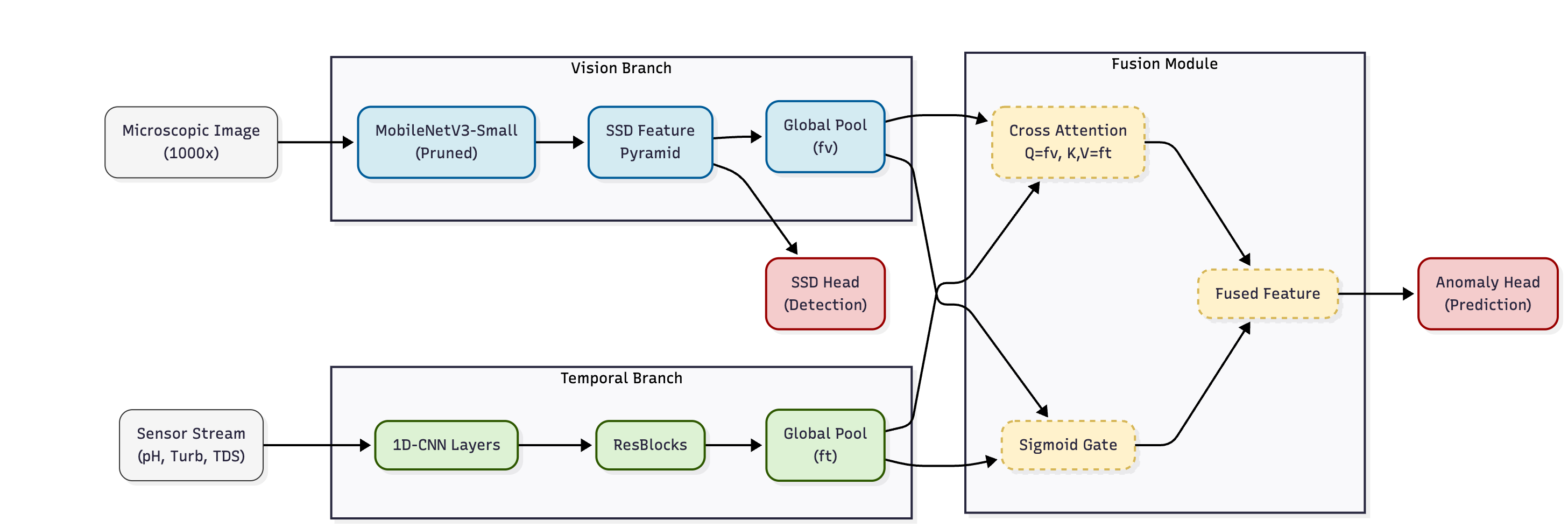}
  \caption{AquaFusionNet architecture. The vision branch uses a pruned SSD--MobileNetV3-Small backbone to detect microorganisms and particulate contaminants in microscopic frames. AquaTemp-Net processes multivariate physicochemical time series (pH, turbidity, TDS, temperature, DO, ORP) into a compact temporal descriptor. A gated cross-attention module fuses visual feature $f_v$ and temporal feature $f_t$ to produce a fused representation $f_{\text{fused}}$ used for anomaly prediction, while the SSD head outputs bounding boxes and class scores.}
  \label{fig:arch}
\end{figure*}

Figure~\ref{fig:arch} provides an overview of the architecture.

\subsubsection{Vision Branch}

The vision branch is based on MobileNetV3-Small~\cite{howard_2019_mobilenetv3} with width multiplier $\alpha = 0.75$ and depthwise separable convolutions for efficiency. We attach an SSD-style feature pyramid~\cite{liu_2016_ssd} that produces multiscale feature maps (P3--P7) for detection of both small and large objects.

To meet the memory and latency budget on Jetson Nano, we perform structured channel pruning guided by layer-wise sensitivity analysis. Channels that contribute least to validation accuracy are iteratively removed, with fine-tuning after each pruning step. We stop pruning when further removal yields more than 0.5\% absolute drop in validation mAP. This procedure eliminates approximately 38\% of channels relative to the unpruned backbone.

The detection head consists of a set of convolutional predictors attached to each pyramid level, outputting class probabilities, bounding box regressions, and objectness scores. Non-maximum suppression (NMS) is applied at inference to obtain the final detections.

\subsubsection{Temporal Branch: AquaTemp-Net}

The temporal branch, AquaTemp-Net, processes 60-second windows of multivariate sensor time series. Let $S_t \in \mathbb{R}^{C \times T}$ denote the sensor matrix at time $t$, where $C=6$ channels (pH, turbidity, TDS, temperature, DO, ORP) and $T$ time steps.

AquaTemp-Net follows a compact residual 1D-CNN design:
\begin{align}
  x_1 &= \text{Conv1D}_{7,64}(S_t) \xrightarrow{\text{BN}} \text{ReLU}, \\
  x_2 &= \text{ResBlock}_{128}(x_1) \times 2, \\
  x_3 &= \text{ResBlock}_{256}(x_2) \times 2, \\
  f_t &= \text{GlobalAvgPool}(x_3),
\end{align}
where $\text{Conv1D}_{k,d}$ denotes a 1D convolution with kernel size $k$ and $d$ output channels, and each residual block has two convolutional layers with identity skip connections. The architecture is intentionally shallow compared to generic time-series models (e.g., large transformers) to keep the parameter count and FLOPs modest.

\subsubsection{Gated Cross-Attention Fusion}

Let $f_v$ denote a global pooled visual feature obtained from the highest-resolution feature map of the vision branch, and $f_t$ the temporal feature from AquaTemp-Net. We first project these into a shared embedding space:
\begin{align}
  q &= W_Q f_v, \quad
  k = W_K f_t, \quad
  v = W_V f_t,
\end{align}
with learnable matrices $W_Q, W_K, W_V \in \mathbb{R}^{d_k \times d}$.

We then compute a cross-attention vector:
\begin{align}
  a &= \text{softmax}\left(\frac{q^\top k}{\sqrt{d_k}}\right) v,
\end{align}
which can be interpreted as a sensor-informed adjustment to the visual feature.

To allow the model to modulate how strongly it trusts the temporal versus visual cues, we introduce a gating mechanism:
\begin{align}
  g &= \sigma\big(W_g [f_v; f_t]\big),\\
  f_{\text{fused}} &= g \odot a + (1-g) \odot f_v,
\end{align}
where $[\cdot;\cdot]$ denotes concatenation, $\odot$ element-wise multiplication, and $W_g$ is a learnable projection. The fused representation $f_{\text{fused}}$ is fed to a small multilayer perceptron (MLP) and a final sigmoid unit that outputs an anomaly risk score in $[0,1]$.

\subsubsection{Inference Pipeline}

Algorithm~\ref{alg:inference} summarises the end-to-end inference procedure at each time step.

\begin{algorithm}[h]
\caption{AquaFusionNet Inference Pipeline}
\label{alg:inference}
\begin{algorithmic}[1]
\REQUIRE Current frame $I_t$, sensor window $S_t$ (60~s)
\STATE $f_v \leftarrow \text{SSD-MobileNetV3}(I_t)$
\STATE $f_t \leftarrow \text{AquaTemp-Net}(S_t)$
\STATE $f_{\text{fused}} \leftarrow \text{GatedCrossAttention}(f_v, f_t)$
\STATE $boxes, scores, classes \leftarrow \text{SSD-Head}(f_v)$
\STATE $risk \leftarrow \text{Sigmoid}(\text{MLP}(f_{\text{fused}}))$
\IF{$risk > \tau_r$ \OR ($\exists$ pathogen with $score > \tau_c$)}
    \STATE Trigger alert and store frame + metadata
\ENDIF
\end{algorithmic}
\end{algorithm}

In our deployment we set $\tau_r = 0.7$ and $\tau_c = 0.6$ based on validation-set calibration; these thresholds can be adjusted for more conservative or liberal alerting depending on operator preferences.

\subsection{Training and Optimization}
\label{subsec:training}

AquaFusionNet is trained using a joint loss:
\begin{align}
  \mathcal{L} = \mathcal{L}_{\text{det}} + \lambda \mathcal{L}_{\text{anom}},
\end{align}
where $\mathcal{L}_{\text{det}}$ is the standard SSD detection loss combining smooth-$L_1$ localization and focal classification terms, and $\mathcal{L}_{\text{anom}}$ is a binary cross-entropy loss on anomaly labels derived from laboratory-confirmed contamination events. The hyperparameter $\lambda$ balances the two; we found $\lambda = 1.0$ to perform well.

We use the Adam optimizer with an initial learning rate of $10^{-3}$, cosine decay, and minibatch size 32. Training is performed in FP32 on a single NVIDIA RTX 4090 GPU for 200 epochs. After convergence, we apply post-training INT8 quantization using calibration on a held-out subset of training data that approximates the distribution of deployment inputs.

Structured pruning is integrated into the training pipeline. We alternate between pruning and fine-tuning phases: in each pruning phase, we remove a small fraction of channels with the lowest importance scores, then fine-tune for several epochs to recover performance. This continues until validation mAP degrades by more than 0.5\% relative to the unpruned baseline.

To facilitate fair comparison, all baseline detectors (YOLOv5n, YOLOv8n, YOLOv10n, PP-PicoDet, RT-DETR-Lite, NanoDet-Plus, DAMO-YOLO, EfficientDet-Lite0) undergo the same INT8 quantization and are evaluated on the same Jetson Nano configuration.

\subsection{Statistical Analysis}
\label{subsec:stats}

We evaluate detection performance using mean Average Precision at IoU threshold 0.5 (mAP@0.5) on the AquaMicro12K test set. For anomaly prediction, we report accuracy, precision, recall, and area under the receiver operating characteristic curve (ROC-AUC). In field deployments, we additionally report false positive rate (FPR) and false negative rate (FNR) of alerts relative to laboratory-confirmed contamination events.

To assess the statistical significance of differences in detection performance between AquaFusionNet and baseline detectors, we use McNemar's test on paired per-image outcomes (correct vs.\ incorrect). For continuous metrics such as mAP and anomaly accuracy across sites, we employ the Wilcoxon signed-rank test to compare AquaFusionNet to the strongest competing model. Significance is assessed at $\alpha = 0.05$, with $p < 0.001$ highlighted where observed.

We also study calibration via reliability diagrams and expected calibration error (ECE) for the anomaly risk scores, though detailed plots are omitted here for brevity.

\section{Results}
\label{sec:results}

We now present quantitative results on the AquaMicro12K benchmark, ablation studies to analyse the contribution of architectural components, and field performance over six months of deployment.

\subsection{Benchmark on AquaMicro12K}
\label{subsec:sota}

Table~\ref{tab:sota} compares AquaFusionNet to representative lightweight detectors on the AquaMicro12K test set. All models are quantized to INT8 and run on Jetson Nano in max-N mode. We report mAP@0.5, parameter count, model size, frames per second (FPS), and measured power draw.

\begin{table*}[t]
\centering
\caption{Comparison on AquaMicro12K test set (INT8 models). Power and FPS measured on Jetson Nano (max-N mode).}
\label{tab:sota}
\begin{tabular}{lccccc}
\toprule
Model & mAP@0.5 & Params (M) & Size (MB) & FPS (Jetson Nano) & Power (W) \\
\midrule
YOLOv5n              & 89.3 & 1.9  & 4.1  & 52 & 5.3 \\
YOLOv8n              & 91.7 & 3.2  & 6.4  & 48 & 5.9 \\
YOLOv10n             & 92.4 & 2.8  & 5.6  & 50 & 6.0 \\
PP-PicoDet-L         & 90.8 & 3.6  & 7.5  & 45 & 5.7 \\
RT-DETR-Lite         & 93.1 & 6.8  & 13.2 & 38 & 6.8 \\
NanoDet-Plus         & 88.9 & 0.95 & 2.1  & 61 & 5.4 \\
DAMO-YOLO            & 92.9 & 4.5  & 9.2  & 42 & 6.3 \\
EfficientDet-Lite0   & 90.5 & 4.0  & 8.1  & 35 & 6.5 \\
\textbf{AquaFusionNet (ours)} & \textbf{94.8} & \textbf{8.7} & \textbf{8.7} & \textbf{41} & \textbf{4.8} \\
\bottomrule
\end{tabular}
\end{table*}

AquaFusionNet achieves the highest mAP@0.5 (94.8\%) among all sub-15\,MB models, while also exhibiting the lowest power consumption. The throughput of 41\,FPS is lower than NanoDet-Plus but sufficient for our 1--2\,FPS acquisition rate, leaving headroom for logging and communication. McNemar tests indicate that improvements over RT-DETR-Lite, the strongest baseline, are statistically significant ($p < 0.001$).

\subsection{Effect of Cross-Modal Fusion}
\label{subsec:ablation}

To quantify the effect of each architectural component, we perform an ablation study on the validation set. Table~\ref{tab:ablation} compares:

\begin{itemize}
  \item A vision-only model (pruned SSD--MobileNetV3) with no temporal branch,
  \item A temporal-only model (AquaTemp-Net) for anomaly prediction,
  \item A simple late-fusion model that concatenates $f_v$ and $f_t$,
  \item A cross-attention fusion model without gating,
  \item The full AquaFusionNet with gating, pruning, and INT8 quantization.
\end{itemize}

\begin{table}[h]
\centering
\caption{Ablation results on validation set. ``Anomaly Acc.'' is the accuracy of sensor-based anomaly prediction.}
\label{tab:ablation}
\begin{tabular}{lcc}
\toprule
Configuration                  & mAP@0.5 & Anomaly Acc. \\
\midrule
Vision only                    & 91.2 & --- \\
Temporal only                  & ---  & 89.7\% \\
Late fusion (concatenation)    & 93.1 & 93.8\% \\
Cross-attention (no gate)      & 94.3 & 95.1\% \\
+ Gating mechanism             & 94.6 & 95.9\% \\
+ INT8 + pruning (full model)  & \textbf{94.8} & \textbf{96.3\%} \\
\bottomrule
\end{tabular}
\end{table}

Simple late fusion already improves both detection and anomaly prediction relative to unimodal baselines, indicating that the two modalities indeed carry complementary information. Cross-attention further refines this combination by allowing sensor features to influence how the visual representation is interpreted. The gating mechanism yields additional gains and stabilises performance in sequences with transient sensor noise or bubbles in the microscopic chamber. Pruning and quantization preserve performance, suggesting that the fused representation is robust to reduced precision.

\subsection{Anomaly Prediction and Calibration}
\label{subsec:calibration}

On the held-out test set, AquaFusionNet achieves 96.3\% anomaly-prediction accuracy, with precision 95.4\%, recall 96.9\%, and ROC-AUC 0.982. The expected calibration error (ECE) of the anomaly risk scores is 0.024 when using temperature scaling learned on the validation set. Reliability diagrams (not shown) indicate that the model tends to be slightly conservative: predicted probabilities for high-risk cases are marginally lower than the empirical frequencies, which is preferable to overconfidence in many safety-critical contexts.

We also examined per-class anomaly contributions. Events associated with \emph{E.~coli} and \emph{Giardia} contamination exhibit strong alignment between microscopic detections and sensor-based risk scores, while microplastic-dominated events are sometimes flagged as lower risk, reflecting current practice where regulations prioritise microbial safety over microplastics.

\subsection{Six-Month Field Performance}
\label{subsec:field}

Across seven deployment sites and six months, AquaFusionNet processed 1.84\,million microscopic frames and their associated sensor windows. Alerts raised by the system were compared against periodic laboratory tests conducted in accordance with national drinking-water standards. Laboratory sampling frequency ranged from weekly to bi-weekly, depending on site.

Figure~\ref{fig:field} summarises the temporal evolution of alerts and laboratory-confirmed contamination events, along with per-site error rates.

\begin{figure}[h]
  \centering
  \includegraphics[width=\columnwidth]{figures/field_results.png}
  \caption{Field performance (January--June 2025). Top: daily number of AquaFusionNet alerts overlaid with laboratory-confirmed contamination events. Bottom: false positive and false negative rates by site, illustrating variability across depots and river intakes.}
  \label{fig:field}
\end{figure}

Aggregated over all sites, the system exhibits:

\begin{itemize}
  \item False positive rate (FPR): 2.3\% of alerts flagged as high risk but not confirmed by laboratory tests.
  \item False negative rate (FNR): 1.1\% of laboratory-confirmed events where AquaFusionNet did not raise a high-risk alert.
  \item Average power consumption: 4.8\,W, enabling continuous operation from a 20\,W solar panel and battery system.
  \item Uptime: 100\% during recorded grid outages, with no missed inference windows due to power loss.
\end{itemize}

When we disable the temporal branch and operate in a vision-only mode, FPR increases to 4.5\%, primarily due to misclassification of bubbles, staining residues, and illumination artefacts as contaminants. In contrast, a sensor-only anomaly model (AquaTemp-Net plus thresholding) yields an FNR of 3.8\% for rare but visually evident contamination events that manifest weakly in the short-horizon sensor window. The cross-modal fusion configuration therefore shifts the error profile toward fewer missed high-risk events while keeping false alarms at a manageable level.

We also observe site-level differences. River intake nodes show both higher alert rates and slightly higher FPR than depots, reflecting genuine variability in raw-water quality and episodes of sensor fouling. These differences suggest that, in practice, site-specific calibration of alert thresholds and maintenance schedules will be necessary.

\section{Discussion}
\label{sec:discussion}

\subsection{Implications for Small-Scale Water Systems}

The empirical results suggest that cross-modal fusion can provide operators of small-scale drinking-water systems with richer, more reliable real-time feedback than unimodal approaches. In particular, the ability to combine microscopic evidence with contemporaneous sensor readings has three practical benefits:

\begin{itemize}
  \item \textbf{Earlier and more specific alerts.} Turbidity spikes that coincide with emerging bacterial clusters are treated differently from spikes associated with harmless inorganic particles. This helps prioritise responses (e.g., immediate tank flushing vs.\ scheduled maintenance).
  \item \textbf{Reduced alarm fatigue.} By filtering out sensor anomalies that are not supported by microscopic evidence, AquaFusionNet reduces false positives relative to threshold-based sensor-only schemes. This may help operators maintain trust in the monitoring system.
  \item \textbf{Context for follow-up action.} Stored micrographs, sensor traces, and risk scores provide a record that can be reviewed by regulators or technical support staff, supporting root-cause analysis and longer-term process improvements.
\end{itemize}

Because the entire inference pipeline runs on an inexpensive edge device, connectivity becomes a convenience rather than a prerequisite. Sites can continue to operate during network outages and synchronise logs when connectivity resumes.

\subsection{Hardware--Software Co-Design}

A key lesson from this work is that multimodal fusion models for edge deployment cannot be designed in isolation from hardware considerations. The specific constraints of Jetson Nano---4\,GB RAM, limited thermal envelope, and power draw---imposed a discipline on architecture design. For example, generic transformers for multimodal fusion were experimentally ruled out early due to their parameter count and inference latency. This pushed us toward gated cross-attention with carefully sized embeddings and a narrow MLP head.

The structured pruning and quantization pipeline is also essential. Without these steps, a straightforward SSD--MobileNetV3 backbone with an unpruned temporal branch would exceed our memory and latency budgets. Co-optimizing model structure, pruning schedules, and quantization calibration allowed us to maintain high accuracy while operating within a realistic power envelope.

These experiences are likely to generalise beyond drinking-water applications, to other environmental monitoring tasks where cameras and sensors coexist on resource-constrained nodes.

\subsection{Limitations}

Several limitations should be acknowledged:

\begin{itemize}
  \item \textbf{Contaminant coverage.} AquaMicro12K focuses on eight classes relevant to the sites studied. Viruses and submicron particles are not captured by our optical setup. Extending the dataset to additional contaminants would require different imaging modalities or staining protocols.
  \item \textbf{Geographical scope.} All field deployments are in East Java. While conditions there are diverse, the generalisability of the model to other regions with different source waters, treatment practices, and operating cultures remains to be validated.
  \item \textbf{Sampling frequency mismatch.} Laboratory tests are conducted weekly or bi-weekly, whereas AquaFusionNet operates continuously. This temporal mismatch may obscure some discrepancies between model alerts and laboratory results. More frequent lab sampling would enable finer-grained evaluation but is operationally costly.
  \item \textbf{Operator integration.} We have not, in this study, systematically evaluated how operators use and respond to AquaFusionNet alerts. Understanding human–AI interaction in this context will be essential for effective deployment.
\end{itemize}

\subsection{Ethical and Governance Considerations}

The introduction of automated decision-support tools into water-safety workflows raises governance questions. While AquaFusionNet can provide early warnings, it should not be viewed as a replacement for regulatory monitoring. Rather, it can complement existing frameworks by offering additional evidence between laboratory tests.

Open-sourcing the dataset, models, and hardware designs is intended to support transparency and independent evaluation. However, care must be taken to ensure that deployments do not exacerbate inequities---for example, if only better-funded depots can afford such systems. Engagement with regulators, local communities, and water providers will be crucial to ensure that the technology is used in ways that are both effective and fair.

\section{Conclusion}
\label{sec:conclusion}

We have presented AquaFusionNet, a lightweight vision--sensor fusion framework for real-time pathogen detection and water-quality anomaly prediction on edge devices. By explicitly modelling the joint evolution of microscopic appearance and short-horizon physicochemical sensor dynamics, AquaFusionNet addresses limitations of unimodal detectors and sensor-only monitoring schemes.

Through the AquaMicro12K dataset, a hardware-aware architecture, and a six-month field deployment across seven drinking-water facilities in Indonesia, we have shown that cross-modal fusion can achieve high detection accuracy, good calibration, and low power consumption on an affordable edge platform. The observed reductions in false positives and false negatives relative to unimodal baselines suggest that such systems can provide more actionable information for operators of small-scale water systems.

Future work will pursue three directions. First, we plan to extend the contaminant taxonomy and incorporate additional imaging modalities to capture a broader range of hazards. Second, we aim to port AquaFusionNet to newer edge accelerators, including NPUs and microcontroller-class devices, to further reduce cost and power consumption. Third, we will work with water utilities and regulators to integrate cross-modal monitoring into operational and governance frameworks, ensuring that algorithmic alerts translate into timely and equitable public-health interventions.

\section*{Acknowledgement}
The authors thank the participating drinking-water depots and local health authorities in Banyuwangi Regency for their collaboration and support during field deployments.

\nocite{*}  

\bibliographystyle{IEEEtran}
\bibliography{references}

\end{document}